\documentclass[11pt]{article}

\usepackage[utf8]{inputenc}
\usepackage[T1]{fontenc}
\usepackage{amsmath,amssymb}
\usepackage{booktabs}
\usepackage{graphicx}
\usepackage{geometry}
\usepackage{hyperref}
\usepackage{xcolor}
\usepackage{authblk}
\usepackage{caption}
\usepackage{tikz}
\usepackage{pgfplots}
\pgfplotsset{compat=1.17}
\usetikzlibrary{positioning,fit,backgrounds,arrows.meta}
\usepgfplotslibrary{fillbetween}

\definecolor{amberfill}{HTML}{FAC775}
\definecolor{amberedge}{HTML}{BA7517}
\definecolor{grayfill}{HTML}{D3D1C7}
\definecolor{grayedge}{HTML}{888780}
\definecolor{tealfill}{HTML}{9FE1CB}
\definecolor{tealedge}{HTML}{0F6E56}
\definecolor{coralfill}{HTML}{F5C4B3}
\definecolor{coraledge}{HTML}{993C1D}
\definecolor{purpfill}{HTML}{CECBF6}
\definecolor{purpedge}{HTML}{534AB7}

\hypersetup{colorlinks=true, linkcolor=blue, citecolor=blue, urlcolor=blue}

\title{\textbf{Grouped Query Experts: Mixture-of-Experts on GQA\\ Self-Attention}}

\author[ ]{Vishesh Tripathi \quad Abhay Kumar}
\affil[ ]{FrontiersMind}
\date{}

\begin{document}
\maketitle

\begin{abstract}
Self-attention is central to Transformer performance and is often the most expensive part of the Transformer at long context lengths, because its pairwise token interactions scale quadratically with sequence length. Standard dense attention also applies the same set of attention heads to every token regardless of token difficulty or information content. This uniform activation can waste compute, especially as sequences grow longer and attention cost increases rapidly. We propose \emph{Grouped Query Experts} (GQE), a mixture-of-experts layer on top of grouped-query attention (GQA): within each GQA group, a router selects $k$ query-head experts per token while all key--value (KV) heads remain dense and unchanged. Thus GQE keeps the KV-cache benefits of GQA and reduces only the active query-head computation. On a fixed 30B-token budget at the 250M-parameter scale, GQE matches the all-active GQA baseline in downstream accuracy while activating half of the routed query-head experts per token, and achieves 1.7--1.8$\times$ prefill speedup at long context lengths.
\end{abstract}

\section{Introduction}

Transformer architectures rely on self-attention to model interactions between tokens~\cite{vaswani2017attention}. This mechanism is powerful but expensive: as sequence length grows, the number of token--token interactions grows quadratically, making self-attention a dominant compute bottleneck in long-context Transformers. Standard attention also spends this compute uniformly: every token activates every attention head. This uniform allocation can be inefficient for heterogeneous token contexts. Content-bearing words, rare terms, and long-range dependencies may need specialized heads, while punctuation, stop words, and other low-information tokens often do not. If the useful heads vary by token, then evaluating all heads for all tokens is unnecessary work.

This paper asks whether the conditional-computation idea behind mixture-of-experts (MoE) models~\cite{shazeer2017outrageously,lepikhin2020gshard,fedus2021switch} can be moved from Transformer MLP blocks into the attention block. We build on grouped-query attention (GQA)~\cite{ainslie2023gqa}, which reduces KV-cache memory and bandwidth by sharing keys and values across groups of query heads~\cite{shazeer2019fast,ainslie2023gqa}. However, GQA still keeps all query heads active. Our method, Grouped Query Experts (GQE), keeps the GQA KV path unchanged and routes each token to $k$ query-head experts within each GQA group. This resizes the output projection $W_O$ from $N \times d$ to $(kG+2) \times d$, a difference we account for in all comparisons.

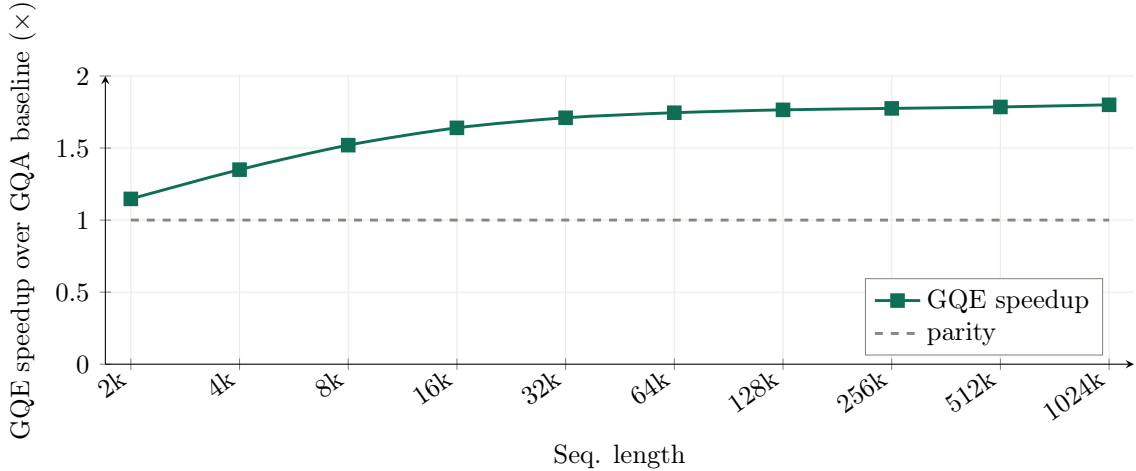
\begin{figure}[t]
\centering
\begin{tikzpicture}
\begin{axis}[
  width=0.92\linewidth, height=5.4cm,
  xlabel={Seq. length},
  ylabel={GQE speedup over GQA baseline ($\times$)},
  xmode=log,
  log basis x=2,
  xtick={2000,4000,8000,16000,32000,64000,128000,256000,512000,1024000},
  xticklabels={2k,4k,8k,16k,32k,64k,128k,256k,512k,1024k},
  ymin=0, ymax=2.0,
  xmin=1700, xmax=1200000,
  legend pos=south east, legend cell align=left,
  legend style={font=\small, draw=grayedge, fill=white},
  tick label style={font=\small}, label style={font=\small},
  axis lines=left, every axis plot/.append style={line width=1.1pt},
  grid=both,
  major grid style={grayfill!45},
  minor grid style={grayfill!25},
  x tick label style={rotate=35, anchor=east},
]
\addplot[tealedge, mark=square*, mark size=2.2pt, smooth] coordinates {
  (2000,1.147)
  (4000,1.350)
  (8000,1.520)
  (16000,1.640)
  (32000,1.710)
  (64000,1.745)
  (128000,1.765)
  (256000,1.775)
  (512000,1.785)
  (1024000,1.800)
};
\addplot[grayedge, dashed] coordinates {
  (2000,1.0)
  (1024000,1.0)
};
\legend{GQE speedup, parity}
\end{axis}
\end{tikzpicture}
\caption{GQE speedup over the GQA baseline, computed as baseline prefill time divided by GQE prefill time. Values above $1\times$ indicate that GQE is faster; the long-context regime shows roughly $1.7$--$1.8\times$ speedup.}
\label{fig:compute}
\end{figure}

Our motivation is to make self-attention more efficient while preserving model quality: if different tokens need different attention heads, then the model should not spend the same query-head compute on every token. The goal is compute reduction \emph{without} quality loss. We do not prune heads permanently, and we do not change the dense KV cache. Instead, we make query-head computation conditional: the model keeps a pool of possible attention patterns, but each token uses only the subset it needs. This distinction is especially relevant for long contexts, where attention cost grows with the quadratic sequence-length term and avoided query-side attention computation becomes increasingly valuable; Figure~\ref{fig:compute} summarizes the resulting measured speedup.

Sparse query-head routing matches the dense GQA baseline only when the router receives a proper learning signal and is stabilized by an always-on shared head.

\paragraph{Contributions.}
\begin{itemize}
  \item We formulate MoE \emph{within} GQA groups: each group contains multiple query-head experts and activates $k$ experts per token, while all KV heads remain dense and always computed. This keeps the comparison aligned with the underlying GQA configuration and preserves GQA's memory profile.
  \item On a fixed 30B-token budget at 250M parameters, we show that GQE matches the all-active GQA baseline in downstream accuracy while activating half of the routed query-head experts per token; including the shared head, the main setting computes 9 of 16 query-attention heads.
  \item We reduce compute in self-attention by activating only a sparse subset of query-head experts per token.
\end{itemize}

\section{Background and Related Work}

\subsection{Mixture of Experts}
MoE layers have become increasingly prominent in modern large language models, where they are most often applied to Transformer feed-forward blocks~\cite{shazeer2017outrageously,lepikhin2020gshard,fedus2021switch}. In an MoE MLP, a router maps each token representation to a small subset of experts; only the selected experts are evaluated, and their outputs are combined using router probabilities. This creates conditional capacity: a model can contain many expert parameters while activating only a fraction for any given token. The key idea is sparsity in the active computation path. Our work transfers this idea from the MLP block to the GQA attention block.

\subsection{Grouped-Query and Multi-Query Attention}
Grouped-query attention (GQA) is an efficient attention variant in which multiple query heads share a smaller number of KV heads~\cite{ainslie2023gqa}. It sits between full multi-head attention (MHA) and multi-query attention (MQA)~\cite{shazeer2019fast}: MHA gives every query head its own key and value head, MQA shares a single KV head across all query heads, and GQA shares one KV head within each query group. For example, a 16-query-head model with 8 KV heads has two query heads per KV group, cutting KV-cache storage relative to 16 full KV heads while retaining more flexibility than a single shared KV head. This is especially useful during autoregressive decoding, where cached keys and values must be stored and read at every generation step.

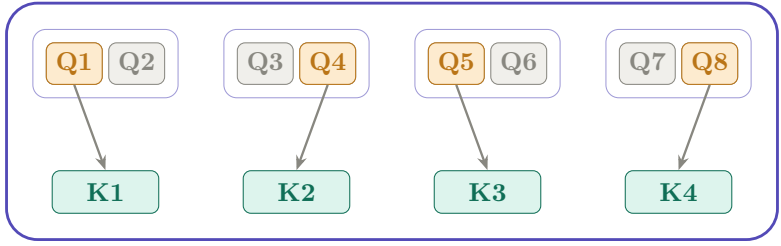
\begin{figure}[t]
\centering
\resizebox{0.62\linewidth}{!}{%
\begin{tikzpicture}[
  qactive/.style={draw=amberedge, fill=amberfill!35, rounded corners=3pt, minimum width=0.72cm, minimum height=0.52cm, font=\small\bfseries, text=amberedge},
  qinactive/.style={draw=grayedge, fill=grayfill!35, rounded corners=3pt, minimum width=0.72cm, minimum height=0.52cm, font=\small\bfseries, text=grayedge},
  kv/.style={draw=tealedge, fill=tealfill!35, rounded corners=3pt, minimum width=1.45cm, minimum height=0.58cm, font=\small\bfseries, text=tealedge},
  grp/.style={draw=purpedge!55, rounded corners=5pt, inner sep=5pt},
  frame/.style={draw=purpedge, rounded corners=12pt, line width=1.1pt, inner sep=10pt},
  ar/.style={-{Stealth[length=2mm]}, grayedge, line width=0.9pt}
]
\foreach \g/\xo/\a/\b/\active in {1/0/1/2/1, 2/2.6/3/4/4, 3/5.2/5/6/5, 4/7.8/7/8/8}{
  \node[qinactive] (q\a) at (\xo,2.7) {Q\a};
  \node[qinactive] (q\b) at (\xo+0.85,2.7) {Q\b};
  \node[qactive] (qa\active) at ({\xo + ifthenelse(\active==\a,0,0.85)},2.7) {Q\active};
  \node[kv] (k\g) at (\xo+0.425,0.95) {K\g};
  \begin{scope}[on background layer]
    \node[grp, fit=(q\a)(q\b)] (grp\g) {};
  \end{scope}
  \draw[ar] (qa\active.south) -- (k\g.north);
}

\begin{scope}[on background layer]
  \node[frame, fit=(grp1)(grp4)(k1)(k4)] {};
\end{scope}
\end{tikzpicture}%
}
\caption{Within-group routing in GQE: each fixed GQA group selects the top-$k$ query experts for a token, while the KV cache remains fixed with four dense KV heads. Highlighted query boxes indicate selected experts; gray boxes indicate inactive experts.}
\label{fig:gqa}
\end{figure}

GQA primarily targets memory bandwidth and KV-cache efficiency; it changes how keys and values are shared, but it still evaluates all query heads for every token. Our method builds directly on this setting: we keep GQA-style KV sharing intact and instead reduce active query-head compute by deciding which query experts inside each GQA group need to run for a given token. This also resizes $W_O$ from $N \times d$ to $(kG+2) \times d$, as the output projection must match the reduced number of active slots. Figure~\ref{fig:gqa} shows the GQA structure our method builds on.

\subsection{Conditional and Sparse Attention Heads}
Several prior works motivate conditional computation inside the attention block. A simple way to view this literature is by asking what is routed, what happens to the KV cache, and whether the routing respects an existing GQA grouping.

Head pruning removes heads after training; for example, if several attention heads are rarely useful, they can sometimes be dropped with limited quality loss~\cite{michel2019sixteen,voita2019analyzing}. Mixture-of-attention methods route over heads or head components; for example, MoA selects a subset of attention heads for each token rather than using every head~\cite{peng2020mixture,zhang2022moa}. SwitchHead and MoH use MoE-style attention modules; for example, MoH gates head outputs while retaining a full per-head KV cache~\cite{csordas2023switchhead,jin2024moh}. Grouped variants route larger head groups; for example, MoMHA forms experts from groups of heads with shared key and value projections, and LLaMA-MoE v2 activates top-$K$ attention-head groups after converting a dense model~\cite{tan2024scattermoe,qu2024llamamoev2}.

\begin{figure}[t]
\centering
\includegraphics[width=0.72\linewidth]{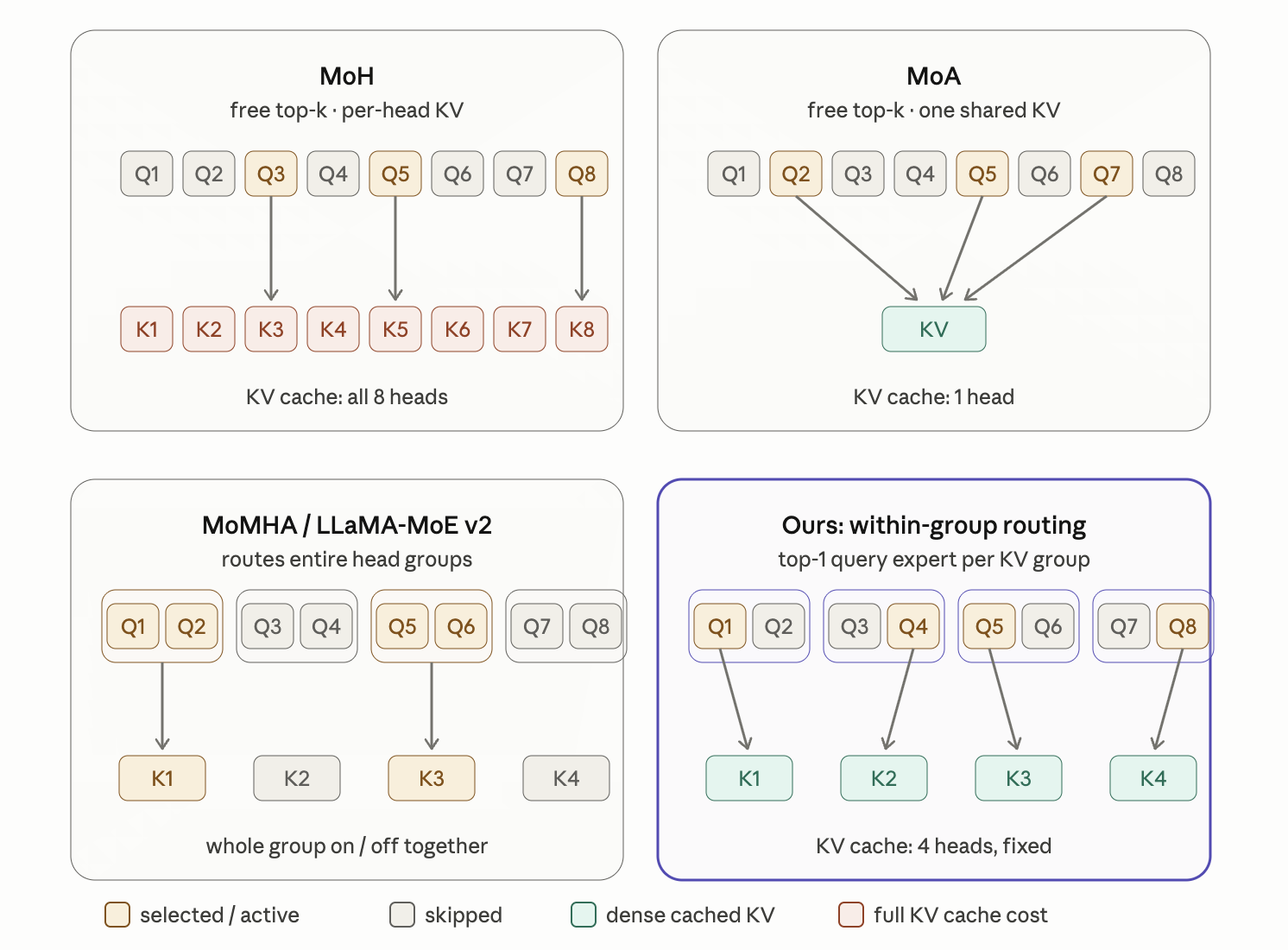}
\caption{Comparison of attention-routing approaches. GQE routes top-$k$ query experts within each fixed GQA group while keeping the KV path dense and unchanged.}
\label{fig:position}
\end{figure}

Our design is simpler and more constrained. \emph{Routing granularity:} instead of choosing top-$k$ heads from the full model, GQE chooses top-$k$ experts inside each fixed GQA group; for example, with 8 KV groups, every group still contributes routed query experts. \emph{KV handling:} instead of changing or sparsifying the KV cache, GQE keeps all GQA KV heads dense; for example, a token may skip some query experts, but it still attends against the same group KV head. \emph{Training-time design:} instead of pruning or converting a trained dense model, GQE trains the router and experts jointly from the start. Figure~\ref{fig:position} contrasts these designs.

\section{Grouped Query Experts}

\subsection{Setup}
Let $X \in \mathbb{R}^{n \times d}$ denote a sequence of $n$ token representations. In standard MHA, every token is projected into queries, keys, and values, and all $H$ heads are evaluated:
\begin{equation}
\mathrm{MHA}(X) = \mathrm{Concat}\big(A_1(X), A_2(X), \dots, A_H(X)\big) W_O,
\end{equation}
where $A_h(\cdot)$ is the computation of head $h$ and $W_O$ is the output projection. In GQA~\cite{ainslie2023gqa}, the $H$ query heads are partitioned into $G$ groups, and all query heads in a group share one KV head. This dense formulation evaluates every query head for every token.

\subsection{Within-Group Query Experts}
We treat the query heads of each GQA group as a set of \emph{experts}. Given a total pool of $N$ query-head experts and $G$ groups (one per KV head), each group contains $M = N/G$ experts $\{E_{g,1}, \dots, E_{g,M}\}$, all of which attend against the same shared KV head of group $g$. Each expert $E_{g,m}$ owns its own query projection and produces an attention-head output; the KV projections are shared within the group and are always computed, so the KV cache is identical to the underlying GQA model. Scaling the expert pool $N$ enlarges the per-group candidate set $M$ without changing the number of routed active experts for fixed $k$ and $G$.

For each routing unit $x_i$ and group $g$, a router produces scores over the $M$ experts. We apply the softmax over the group's experts \emph{before} selection,
\begin{equation}
p_{i,g} = \mathrm{softmax}\big(r_g(x_i)\big) \in \mathbb{R}^M,
\end{equation}
and then select the top-$k$ highest-scoring experts in the group,
\begin{equation}
\mathcal{K}_{g}(x_i) = \operatorname{TopK}_{m}\big(p_{i,g,m}, k\big).
\end{equation}
Applying the softmax before top-$k$ selection gives per-group probabilities used to decide which expert is selected from each group. For example, with eight query experts split into four groups, $\{a_1,a_2\}$, $\{a_3,a_4\}$, $\{a_5,a_6\}$, and $\{a_7,a_8\}$, the router may select $a_1,a_3,a_5,a_7$ if those experts have the larger probabilities within their respective groups.

\subsection{Output Construction with a Shared Head}
Let $o_{i,g,m} = E_{g,m}(x_i)$ be the output of selected expert $m \in \mathcal{K}_{g}(x_i)$ in group $g$, and let $s(x_i)$ be an always-on shared attention head that is computed for every token regardless of routing. The selected expert outputs are first kept as ordinary, unscaled head slots. In the example above, the hard-routed part is the concatenation of the selected heads, e.g., $\mathrm{Concat}(a_1,a_3,a_5,a_7)$, not a weighted average:
\begin{equation}
O_i = \big\{o_{i,g,m} : g \in \{1,\dots,G\},\; m \in \mathcal{K}_{g}(x_i)\big\}.
\end{equation}
The selected probabilities from different groups do not generally sum to one after top-$k$ selection. We therefore add one extra router-supervised slot: a renormalized weighted sum of the selected expert outputs,
\begin{equation}
\bar{o}_i = \sum_{g=1}^{G} \sum_{m \in \mathcal{K}_{g}(x_i)} w_{i,g,m}\, o_{i,g,m},
\qquad
w_{i,g,m} = \frac{p_{i,g,m}}{\sum_{g'=1}^{G} \sum_{m' \in \mathcal{K}_{g'}(x_i)} p_{i,g',m'}}.
\end{equation}
\paragraph{Router learning signal.}
The selected router probabilities are used only in the weighted-sum slot, which provides a differentiable path from the language-modeling loss back to the router. Without this weighted-sum slot, the hard concatenation uses the selected expert outputs but gives the router a much weaker learning signal, because the discrete top-$k$ decision itself is not differentiable.

The final attention output concatenates the hard-selected expert slots, the weighted-sum slot, and the shared-head slot before the output projection:
\begin{equation}
y_i = \mathrm{Concat}\Big(O_i,\; \bar{o}_i,\; s(x_i)\Big)\, W_O.
\end{equation}
Thus, in the 16-query-head / 8-KV-head setting used in our experiments with $k=1$, the router selects 8 query experts, producing 8 hard-concatenated expert attention outputs; these are concatenated with one renormalized weighted-sum output and one shared attention output, giving 10 slots before $W_O$. This changes the output-projection shape relative to the dense baseline: the baseline projects 16 head slots, whereas this GQE variant projects 10 slots, so $W_O$ is reshaped from a $16d_h \times d$ input projection to a $10d_h \times d$ input projection. Consequently, the comparison keeps the training budget, data, KV layout, and per-head dimension fixed, but it is not exactly parameter-matched in the attention output projection. For general $k$, the routed expert slots scale to $kG$. The expert slots remain unscaled. Only the weighted-sum slot uses router weights, and this slot is the path through which the language-modeling loss gives gradient signal to the router. The always-on shared head $s(\cdot)$ provides a stable, token-independent attention pathway that anchors training while the routed experts specialize.

\subsection{Routing Auxiliary Loss}
In addition to the language-modeling loss, we use a standard load-balancing auxiliary loss to prevent the router from collapsing onto the same expert in each group~\cite{shazeer2017outrageously,fedus2021switch}. The auxiliary loss is computed from the router probabilities and the selected experts within each group: it encourages high-probability experts to be used while also keeping the selected-head distribution balanced across the expert pool. In the $k=1$ case, routing chooses the highest-probability expert from each group for each token, and the auxiliary loss encourages those selections to remain balanced across tokens rather than repeatedly selecting the same head. This auxiliary objective determines the routing behavior together with the language-modeling gradient that flows through the renormalized weighted-sum slot.

\subsection{Compute Profile}
With $k$ active experts per group, exactly $kG$ routed query experts are active per token, out of a total routed pool of $N = MG$ experts. The routed-only active fraction is $kG/N = k/M$. The always-on shared head adds one additional query-attention computation, while the weighted-sum slot reuses the selected expert outputs and does not require an additional attention computation. Therefore the total active query-attention fraction relative to the $N$ routed expert pool is
\begin{equation}
\frac{kG + 1}{N}.
\end{equation}
In the main 16-query-head / 8-KV-head setting with $k=1$, this is $(8+1)/16 = 9/16 \approx 56\%$: the ``50\%'' figure refers only to the routed experts, not to total active query-attention operations. A larger expert pool $M$ lowers the routed active fraction for fixed $k$, while increasing $k$ trades more active compute for more per-token attention capacity. The idealized active-compute reduction is proportional to this sparse fraction, though realized speedup also depends on router overhead, dispatch efficiency, hardware utilization, sequence length, and the smaller $W_O$ input described above. Because the method is applied on top of GQA and leaves the KV path dense, the additional saving beyond GQA is \emph{not} KV-cache memory; it is reduced active query-head computation from skipping unneeded experts. This saving increases with sequence length because attention includes sequence-length-dependent computation~\cite{vaswani2017attention}, which is reflected in the long-context speedups in Figure~\ref{fig:compute}.

\section{Experiments and Results}

\subsection{Experimental Setup}
All ablations are trained on a fixed budget of 30B tokens at the 250M-parameter scale. We use a 30B-token sample from FineWeb-Edu, which is a subset of FineWeb2~\cite{penedo2025fineweb2}. This controlled-token setting ensures that differences between models reflect architectural and routing choices rather than differences in training exposure. We compare GQA baselines with 8 KV heads against GQE variants that add routing over the corresponding GQA query heads. Since the number of groups equals the number of KV heads, the 8-KV-head baseline uses an 8-group layout; in the reported $k=1$ experiments, each group activates one expert and the query heads of the dense baseline become the per-group expert pool. Unless otherwise stated, the tokenizer, optimizer, learning-rate schedule, batch size, sequence length, and training data mixture are held fixed across runs, and the per-head dimension is held constant so that comparisons isolate routing rather than head width. We include training-loss plots in Appendix~\ref{app:loss-graphs}. The training hyperparameters are mentioned in Table~\ref{tab:hyperparams}. We employ ZClip~\cite{kumar2025zclip} to mitigate loss spikes.

\begin{table}[h]
\centering
\begin{tabular}{ll}
\toprule
\textbf{Hyperparameter} & \textbf{Value} \\
\midrule
Optimizer & Fused AdamW ($\beta_1 = 0.9$, $\beta_2 = 0.95$, $\epsilon = 1 \times 10^{-7}$) \\
Learning-rate schedule & WSD \\
Maximum learning rate & $1 \times 10^{-5}$ to $5 \times 10^{-4}$ \\
Warm-up tokens & 3 billion tokens (3BT) \\
Weight decay & 0.1 (AdamW implementation) \\
Global batch size & 1.05 million tokens \\
Sequence length & 2048 \\
Precision & Mixed precision BFloat16 \\
\bottomrule
\end{tabular}
\caption{Training hyperparameters used for the main experiments}
\label{tab:hyperparams}
\end{table}

We report downstream quality on HellaSwag~\cite{zellers2019hellaswag}, PIQA~\cite{bisk2020piqa}, and ARC-Easy~\cite{clark2018arc} rather than a single task, and we report throughput as context length increases in Figure~\ref{fig:compute}, since long contexts are where attention compute dominates and where sparse query-head activation should show the clearest benefit.

\subsection{Accuracy}
Table~\ref{tab:ablation} establishes the headline accuracy result: at the same training budget, the corrected GQE variant matches the all-active GQA baseline within $0.2$ average points while activating half of the routed query-head experts, or 9 of 16 total query-attention computations after including the always-on shared head. Additional downstream task plots are provided in Appendix~\ref{app:downstream-graphs}. Our claim is therefore that GQE \emph{matches its GQA baseline at reduced active query-attention compute}, with the caveat that the GQE output projection has fewer input slots than the dense baseline. The two degraded variants confirm that this match depends on the routing fixes rather than on sparsity alone.

A natural expectation is that any reasonable sparse routing scheme will preserve quality, since the model retains the same heads and merely chooses among them. We find this is not the case. Table~\ref{tab:ablation} reports an ablation ladder at the 16-query-head / 8-KV-head configuration, all trained on 30B tokens with $k=1$ active expert per group.

\begin{table}[h]
\centering
\begin{tabular}{lcccc}
\toprule
\textbf{Variant} & \textbf{HellaSwag} & \textbf{ARC-E} & \textbf{PIQA} & \textbf{Average} \\
\midrule
GQA baseline (all 16 heads active) & \textbf{41.31} & 61.36 & 64.90 & 55.86 \\
\midrule
Weighted concat, no renormalized slot & 40.16 & 60.52 & 64.85 & 55.18 \\
Hard concat only & 40.66 & 60.56 & \textbf{65.07} & 55.43 \\
\textbf{GQE (renorm.\ scoring + shared head)} & 41.01 & \textbf{62.41} & 64.69 & \textbf{56.04} \\
\bottomrule
\end{tabular}
\caption{Ablation over routing and output-construction choices at 16 query heads / 8 KV heads with $k=1$ active expert per group, trained on 30B tokens. Average is over HellaSwag, ARC-Easy, and PIQA.}
\label{tab:ablation}
\end{table}

The ladder tells a clear causal story. A weighted-concat variant without the renormalized router-supervised slot reaches only 55.18 average, $0.68$ below the dense baseline. Hard-concatenating the selected expert outputs improves on this (55.43) but still sits $0.43$ below baseline, because the routed expert slots themselves do not provide a differentiable router path. Adding the renormalized weighted-sum slot together with an always-on shared head reaches 56.04, recovering---and marginally exceeding---the all-active GQA baseline at 55.86, while activating only 8 of 16 routed query heads per token and 9 of 16 query-attention computations in total after including the shared head. The interpretation is that sparse query-head routing succeeds only when (i) the router receives a proper, normalized learning signal through the weighted-sum slot so it can learn \emph{which} expert each token needs, and (ii) a stable shared pathway anchors the layer while the routed experts specialize. Absent either, sparsity costs accuracy.

\subsection{Throughput}
\label{sec:throughput}
The purpose of the method is compute reduction, so throughput is a primary result. Figure~\ref{fig:compute} reports measured prefill speedup, computed as GQA baseline latency divided by GQE latency, across sequence lengths from 2k to 1024k tokens. A value above $1\times$ therefore means that GQE is faster. At 2k tokens, the speedup is modest ($1.15\times$), where routing and dispatch overheads are relatively large compared with the attention work being saved. From 4k tokens onward, GQE consistently reaches the long-context regime, with measured speedups between roughly $1.67\times$ and $1.80\times$. This trend matches the design goal: GQE keeps the KV path dense but skips inactive query experts, so the benefit becomes clearer as sequence length increases and query-side attention work dominates fixed routing overhead.

\section{Limitations \& Future Work}
The results are at the 250M-parameter scale and a 30B-token budget; the small accuracy margin over the baseline should be confirmed with multiple seeds and at larger scales before being treated as robust, and we report it as a match rather than an improvement. Future work will compare GQE against alternative long-context architectures such as Mamba and evaluate whether the same routing benefits hold in larger Transformer architectures. Finally, our main experiments use a small per-group expert pool, so the per-group routing headroom is limited; larger expert pools ($M = N/G$) offer more candidates per group and potentially more specialization, but a broad sweep over $N$ was not the focus here.

\section{Conclusion}
We present MoE on top of GQA self-attention as a compute-reduction method that applies conditional computation to GQA query heads while leaving the KV path dense. Instead of evaluating all GQA query heads for every token, the model routes each token to a sparse subset of query-head experts within its GQA groups. We show that this preserves a large set of possible attention patterns while reducing active computation, but only when the router is given a proper learning signal through renormalized scoring and anchored by an always-on shared head: straightforward routing falls below the baseline, and the corrected method recovers it. On a 30B-token budget, GQE matches the corresponding GQA baseline in downstream accuracy while activating half of the routed query-head experts, corresponding to 9 of 16 total query-attention computations in the main setting once the shared head is included.

\section*{Appendix}
\addcontentsline{toc}{section}{Appendix}
\renewcommand{\thesubsection}{\Alph{subsection}}

\subsection{Loss Graphs}
\label{app:loss-graphs}
\begin{figure}[h]
\centering
\includegraphics[width=0.92\linewidth, trim=0 0 8pt 0, clip]{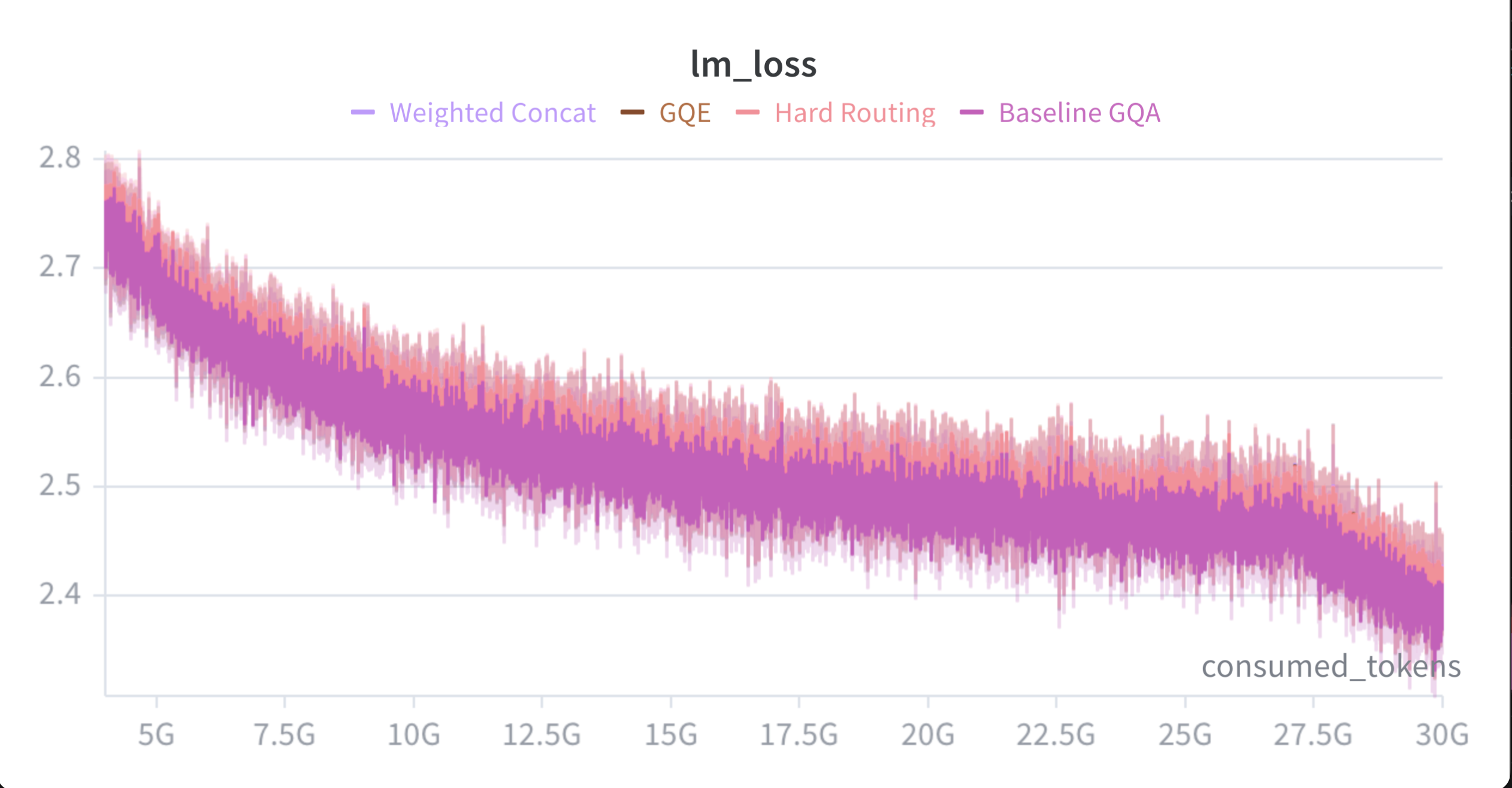}
\caption{Training-loss curves for the four Table~\ref{tab:ablation} variants.}
\label{fig:appendix-loss-demo}
\end{figure}

Figure~\ref{fig:appendix-loss-demo} provides the training-loss curves for the four variants compared in Table~\ref{tab:ablation}.

\subsection{Downstream Task Accuracy Graphs}
\label{app:downstream-graphs}
\begin{figure}[h]
\centering
\includegraphics[width=0.92\linewidth]{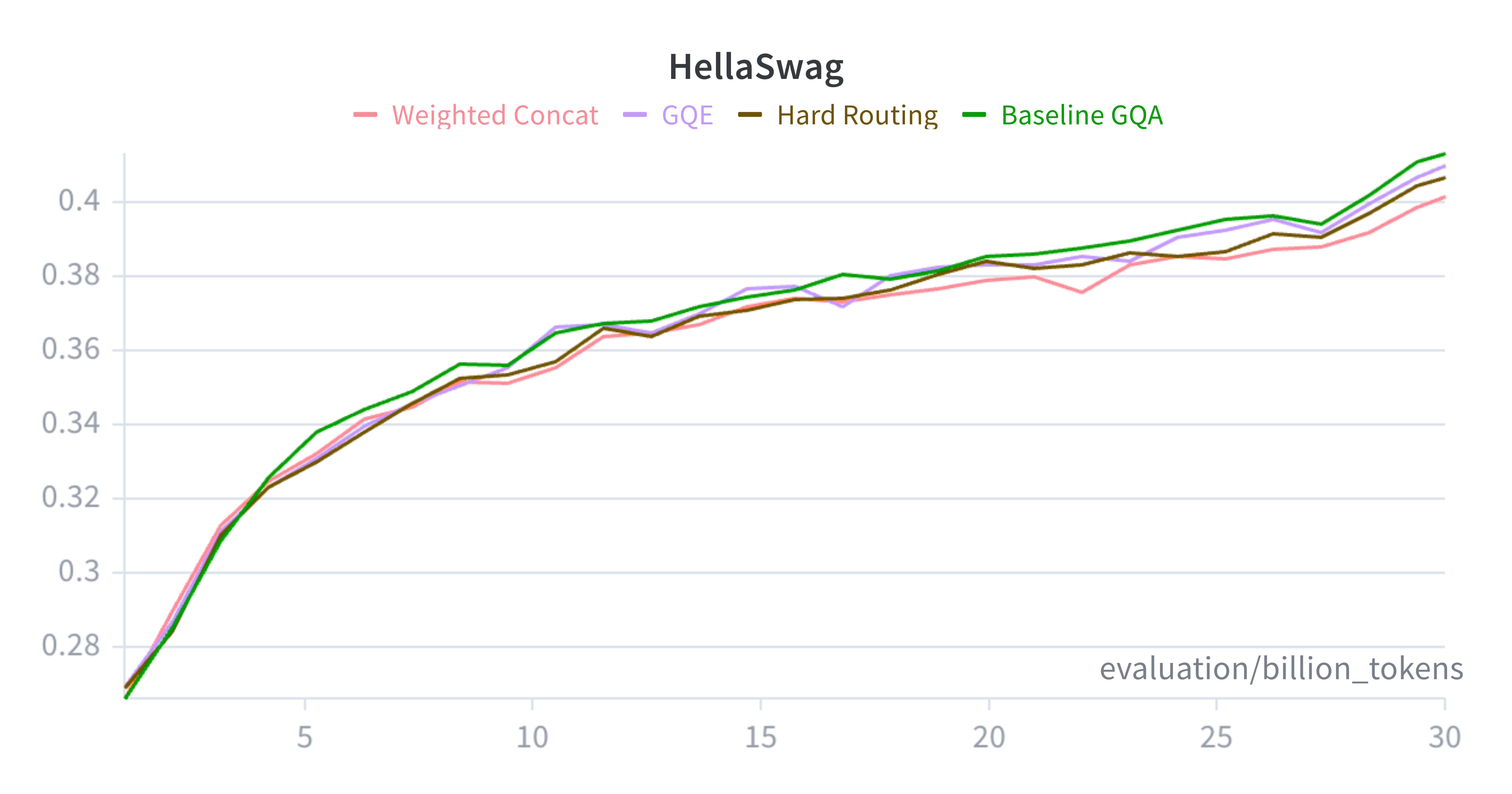}
\caption{HellaSwag accuracy over training tokens for the GQA baseline, routing ablations, and final GQE model.}
\label{fig:appendix-hellaswag-demo}
\end{figure}

\begin{figure}[h]
\centering
\includegraphics[width=0.92\linewidth]{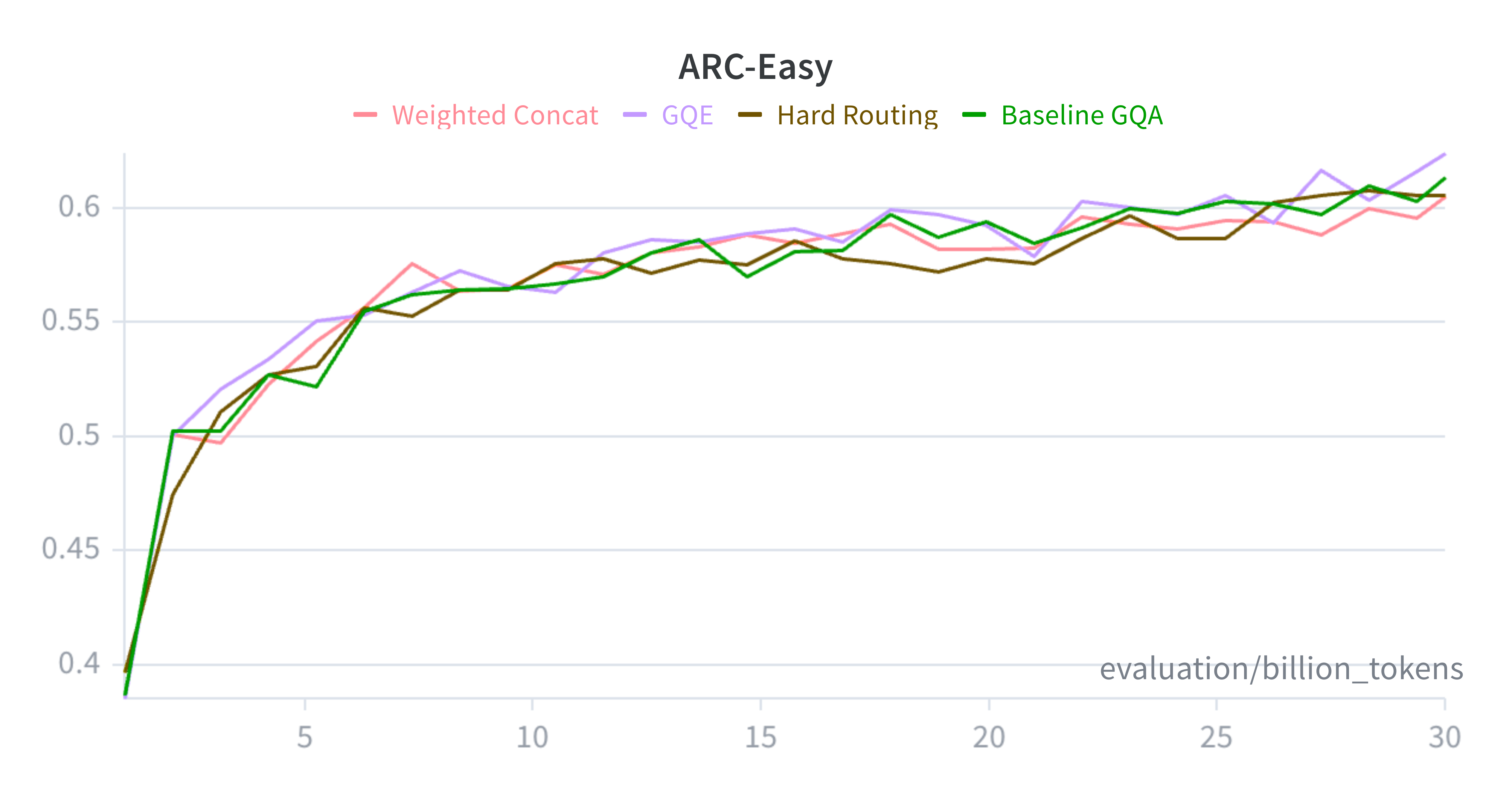}
\caption{ARC-Easy accuracy over training tokens for the GQA baseline, routing ablations, and final GQE model.}
\label{fig:appendix-arce-demo}
\end{figure}

\begin{figure}[h]
\centering
\includegraphics[width=0.92\linewidth]{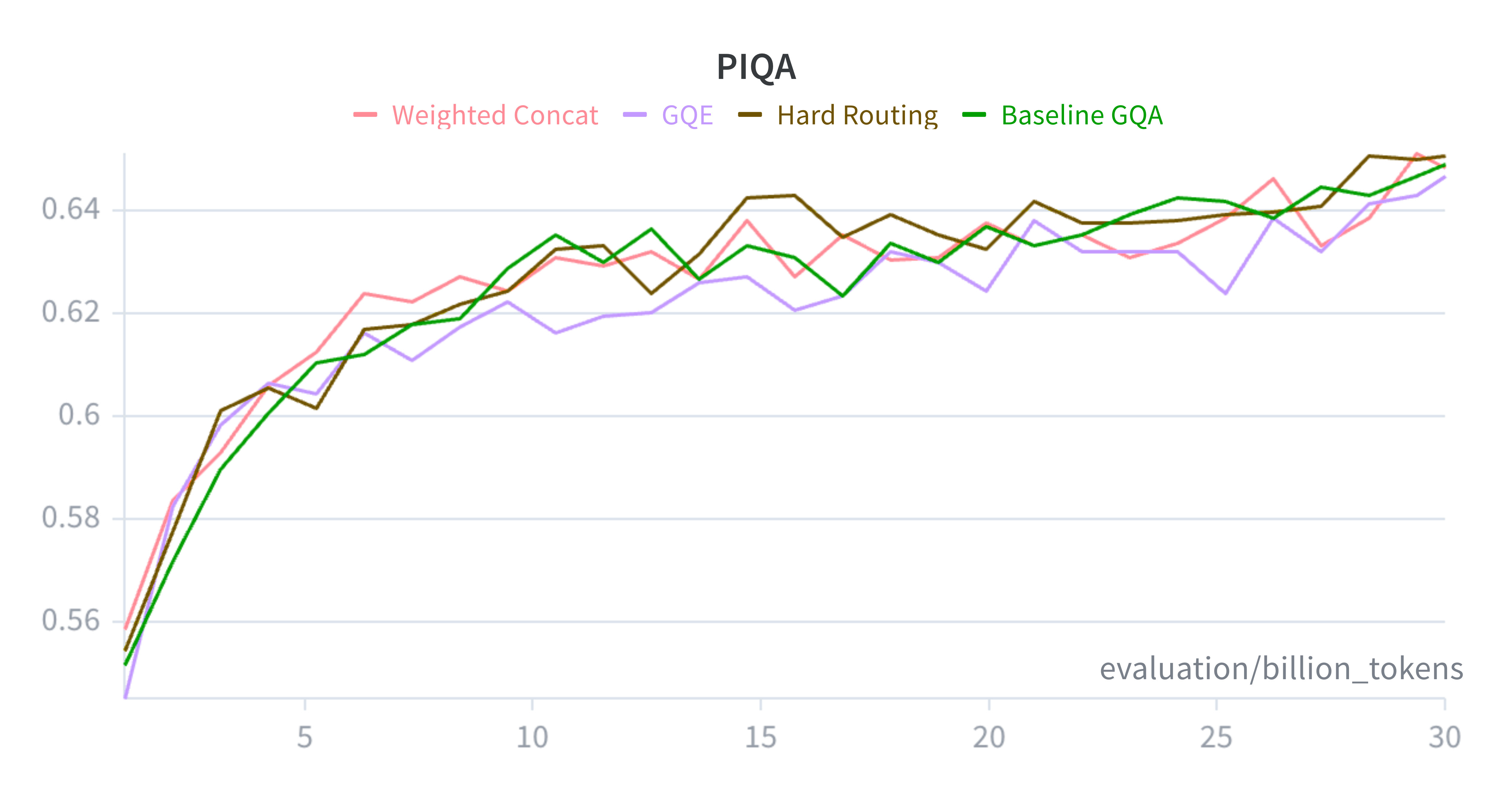}
\caption{PIQA accuracy over training tokens for the GQA baseline, routing ablations, and final GQE model.}
\label{fig:appendix-piqa-demo}
\end{figure}

Figures~\ref{fig:appendix-hellaswag-demo}--\ref{fig:appendix-piqa-demo} show downstream accuracy trends for HellaSwag, ARC-Easy, and PIQA over the full 30B-token training budget. The curves compare the GQA baseline, the intermediate routing ablations, and the final GQE configuration reported in Table~\ref{tab:ablation}.

\end{document}